\documentclass[10pt,twocolumn,letterpaper]{article}

\usepackage{cvpr}
\usepackage{times}
\usepackage{epsfig}
\usepackage{graphicx}
\usepackage{amsmath}
\usepackage{color}
\usepackage{mathtools}
\usepackage{epstopdf}
\usepackage{amssymb}
\usepackage{leftidx}
\usepackage{pifont}
\usepackage{multirow}
\usepackage{makecell}

\usepackage[linesnumbered,ruled]{algorithm2e}
\usepackage{algpseudocode}

\newcommand{\tabincell}[2]{\begin{tabular}{@{}#1@{}}#2\end{tabular}}

\newcommand{\PreserveBackslash}[1]{\let\temp=\\#1\let\\=\temp}
\newcolumntype{C}[1]{>{\PreserveBackslash\centering}p{#1}}
\newcolumntype{R}[1]{>{\PreserveBackslash\raggedleft}p{#1}}
\newcolumntype{L}[1]{>{\PreserveBackslash\raggedright}p{#1}}

\usepackage[pagebackref=true,breaklinks=true,letterpaper=true,colorlinks,bookmarks=false]{hyperref}

\cvprfinalcopy 

\def\cvprPaperID{3821} 

\ifcvprfinal\fi
\begin{document}

\title{Siamese Cascaded Region Proposal Networks for Real-Time Visual Tracking}

\author{Heng Fan \;\;\;\;\; Haibin Ling\\
Department of Computer and Information Sciences, Temple University\\
{\tt\small \{hengfan,hbling\}temple.edu}
}

\maketitle
\thispagestyle{empty}

\begin{abstract}
   Region proposal networks (RPN) have been recently combined with the Siamese network for tracking, and shown excellent accuracy with high efficiency. Nevertheless, previously proposed one-stage Siamese-RPN trackers degenerate in presence of similar distractors and large scale variation. Addressing these issues, we propose a multi-stage tracking framework, Siamese Cascaded RPN (C-RPN), which consists of a sequence of RPNs cascaded from deep high-level to shallow low-level layers in a Siamese network. Compared to previous solutions, C-RPN has several advantages: {\bf (1)} Each RPN is trained using the outputs of RPN in the previous stage. Such process stimulates hard negative sampling, resulting in more balanced training samples. Consequently, the RPNs are sequentially more discriminative in distinguishing difficult background (\ie, similar distractors). {\bf (2)} Multi-level features are fully leveraged through a novel feature transfer block (FTB) for each RPN, further improving the discriminability of C-RPN using both high-level semantic and low-level spatial information. {\bf (3)} With multiple steps of regressions, C-RPN progressively refines the location and shape of the target in each RPN with adjusted anchor boxes in the previous stage, which makes localization more accurate. C-RPN is trained end-to-end with the multi-task loss function. In inference, C-RPN is deployed as it is, without any temporal adaption, for real-time tracking. In extensive experiments on OTB-2013, OTB-2015, VOT-2016, VOT-2017, LaSOT and TrackingNet, C-RPN consistently achieves state-of-the-art results and runs in real-time.

\end{abstract}

\section{Introduction}

\begin{figure}
	\centering
	\includegraphics[width=0.47\linewidth]{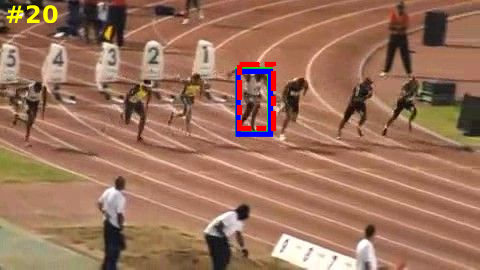} \includegraphics[width=0.47\linewidth]{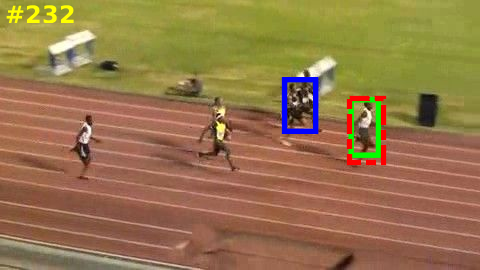} \\
	\vspace{0.1em}
	\includegraphics[width=0.47\linewidth]{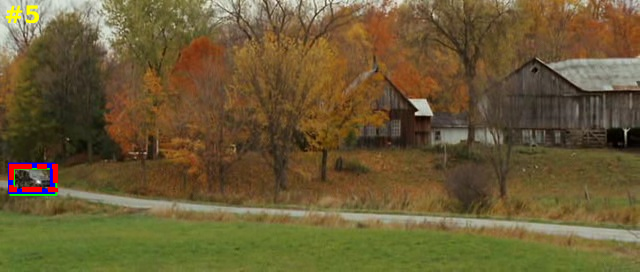} \includegraphics[width=0.47\linewidth]{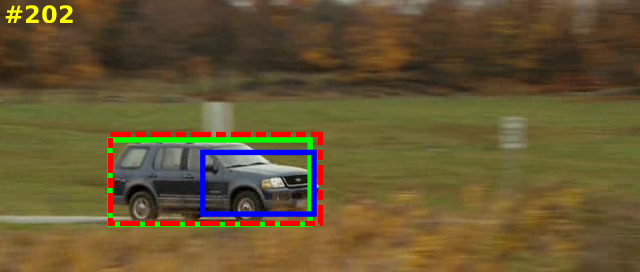} \\
	\includegraphics[width=0.8\linewidth]{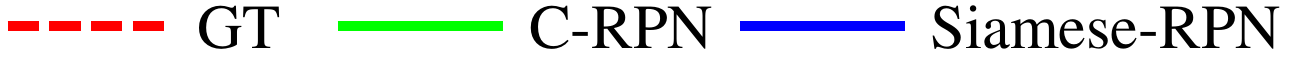} \\
	\caption{Comparisons between one-stage Siamese-RPN~\cite{li2018high} and C-RPN on two challenging sequences: {\it Bolt2} (the top row) with similar distractors and {\it CarScale} (the bottom row) with large scale changes. We observe that C-RPN can distinguish the target from distractors, while Siamese-RPN drifts to the background in {\it Bolt2}. In addition, compared to using a single regressor in Siamese-RPN, multi-regression in C-RPN can better localize the target in presence of large scale changes in {\it CarScale}. Best viewed in color.}
	\label{fig:1}
\vspace{-2mm}
\end{figure}

Visual tracking is one of the most fundamental problems in computer vision, and has a long list of applications such as robotics, human-machine interaction, intelligent vehicle, surveillance and so forth. Despite great advances in recent years, visual tracking remains challenging due to many factor including occlusion, scale variation, etc.

Recently, Siamese network has drawn great attention in the tracking community owing to its balanced accuracy and speed. By formulating object tracking as a matching problem, Siamese trackers~\cite{tao2016siamese,bertinetto2016fully,valmadre2017end,he2018twofold,held2016learning,li2018high,wang2018learning,zhu2018distractor} aim to learn {\it offline} a generic similarity function from a large set of videos. Among these methods, the work of~\cite{li2018high} proposes a one-stage Siamese-RPN for tracking by introducing the regional proposal network (RPN), originally used for object detection~\cite{ren2015faster,liu2016ssd}, into Siamese network. With the proposal extraction by RPN, this approach simultaneously performs classification and localization from multiple scales, achieving excellent performance. Besides, the use of RPN avoids applying the time-consuming pyramid for target scale estimation~\cite{bertinetto2016fully}, resulting in a super real-time solution.

\subsection{Problem and Motivation}

Despite having achieved promising result, Siamese-RPN may drift to the background especially in presence of similar semantic distractors (see Fig.~\ref{fig:1}). We identify two reasons accounting for this.

First, the distribution of training samples is imbalanced: (1) positive samples are far less than negative samples, leading to ineffective training of the Siamese network; and (2) most negative samples are easy negatives (non-similar non-semantic background) that contribute {\it little} useful information in learning a discriminative classifier~\cite{lin2017focal}. As a consequence, the classifier is dominated by the easily classified background samples, and degrades when encountering difficult similar semantic distractors.

Second, low-level spatial features are not fully explored. In Siamese-RPN (and other Siamese trackers), only features of the last layer, which contain more semantic information, are explored to distinguish target/background. In tracking, nevertheless, background distractors and the target may belong to the same category, and/or have similar semantic features~\cite{wang2015visual}. In such case, the high-level semantic features are less discriminative in distinguishing target/background.

In addition to the issues above, the one-stage Siamese-RPN applies a single regressor for target localization using pre-defined anchor boxes. These boxes are expected to work well when having a high overlap with the target. However, for {\it model-free} visual tracking, no prior information regarding the target object is known, and it is hard to estimate how the scale of target changes. Using pre-defined coarse anchor boxes in a single step regression is insufficient for accurate localization~\cite{gidaris2015object,cai2018cascade} (see again Fig.~\ref{fig:1}).

The class imbalance problem is addressed in two-stage object detector (\eg, Faster R-CNN~\cite{ren2015faster}). The first proposal stage rapidly filters out most background samples, and then the second classification stage adopts sampling heuristics such as a fixed foreground-to-background ratio to maintain a manageable balance between foreground and background. In addition, two steps of regressions achieve accurate localization even for objects with extreme shapes.

Motivated by the two-stage detector, we propose a multi-stage tracking framework by cascading a sequence of RPNs to solve the class imbalance problem, and meanwhile fully explore features across layers for robust visual tracking.

\subsection{Contribution}

As the {\bf first contribution}, we present a novel multi-stage tracking framework, the Siamese Cascaded RPN (C-RPN), to solve the problem of class imbalance by performing hard negative sampling~\cite{viola2001rapid,shrivastava2016training}. C-RPN consists of a sequence of RPNs cascaded from the high-level to the low-level layers in the Siamese network. In each stage (level), an RPN performs classification and localization, and outputs the classification scores and the regression offsets for the anchor boxes in this stage. The easy negative anchors are then filtered out, and the rest, treated as hard examples, are utilized as training samples for the RPN of the next stage. Through such process, C-RPN performs stage by stage hard negative sampling. As a result, the distributions of training samples are sequentially more balanced, and the classifiers of RPNs are sequentially more discriminative in distinguishing more difficult distractors (see Fig.~\ref{fig:1}).

{\bf Another benefit} of C-RPN is the more accurate target localization compared to the one-stage SiamRPN~\cite{li2018high}. Instead of using the pre-defined coarse anchor boxes in a single regression step, C-RPN consists of multiple steps of regressions due to multiple RPNs. In each stage, the anchor boxes (including {\it locations} and {\it sizes}) are adjusted by the regressor, which provides better initialization for the regressor of next stage. As a consequence, C-RPN progressively refines the target bounding box, leading to better localization as shown in Fig.~\ref{fig:1}.

\begin{figure*}
	\centering
	\includegraphics[width=1\linewidth]{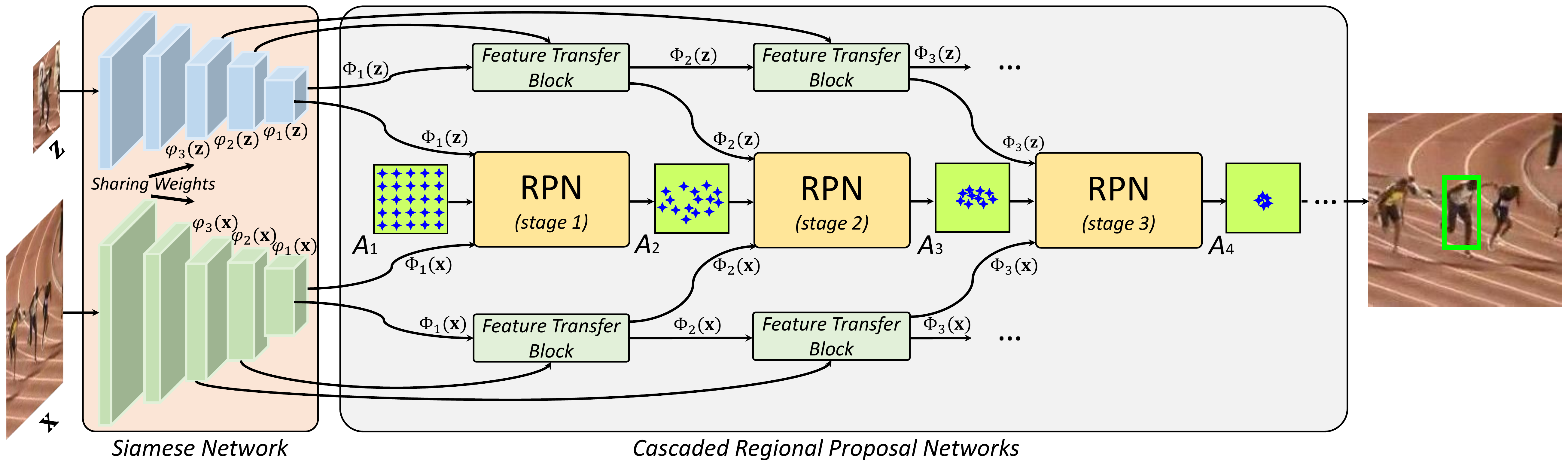}
	\caption{Illustration of the architecture of C-RPN, including a Siamese network for feature extraction and cascaded regional proposal networks for sequential classifications and regressions. The FTB transfers the high-level semantic features for the low-level RPN, and ``A'' represents the set of anchor boxes, which are gradually refined stage by stage. Best viewed in color.}
	\label{fig:cpn}
\vspace{-2mm}
\end{figure*}

Leverage features from different layers in the neural networks has been proven to be beneficial for improving model discriminability~\cite{long2015fully,lin2017refinenet,lin2017feature}. To fully explore both the high-level semantic and the low-level spatial features for visual tracking, we make the {\bf second contribution} by designating a novel feature transfer block (FTB). Instead of separately using features from a single layer in one RPN, FTB enables us to fuse the high-level features into low-level RPN, which further improves its discriminative power to deal with complex background, resulting in better performance of C-RPN. Fig.~\ref{fig:cpn} illustrates the framework of C-RPN.

Last but not least, the {\bf third contribution} is to implement a tracker based on the proposed C-RPN. In extensive experiments on six benchmarks, including OTB-2013~\cite{wu2013online}, OTB-2015~\cite{wu2015object}, VOT-2016~\cite{kristan2016visual},  VOT-2017~\cite{kristan2017visual}, LaSOT~\cite{fan2018lasot} and TrackingNet~\cite{muller2018trackingnet}, our C-RPN consistently achieves the state-of-the-art results and runs in real-time.

\section{Related Work}

Visual tracking has been extensively researched in recent decades. In the following we discuss the most related work, and refer readers to~\cite{smeulders2013visual,yilmaz2006object,li2018deep} for recent surveys.

\vspace{0.12em}
\noindent{\bf Deep tracking.} Inspired by the successes in image classification~\cite{krizhevsky2012imagenet,he2016deep}, deep convolutional neural network (CNN) has been introduced into visual tracking and demonstrated excellent performances~\cite{wang2013learning,wang2015visual,nam2016learning,danelljan2017eco,fan2017sanet,ma2015hierarchical,danelljan2016beyond,song2018vital}. Wang {\it et al.}~\cite{wang2013learning} propose a stacked denoising autoencoder to learn generic feature representation for object appearance modeling in tracking. Wang {\it et al.}~\cite{wang2015visual} introduce a fully convolutional neural network tracking (FCNT) approach by transferring the pre-trained deep features to improve tracking accuracy. Ma {\it et al.}~\cite{ma2015hierarchical} replace hand-craft features in correlation filter tracking with deep features, achieving remarkable gains. Nam and Han~\cite{nam2016learning} propose a light architecture of CNNs with online fine-tuning to learn generic feature for tracking target. Fan and Ling~\cite{fan2017sanet} extend this approach by introducing  a recurrent neural network (RNN) to capture object structure. Song {\it et al.}~\cite{song2018vital} apply adversary learning in CNN to learn richer representation for tracking. Danelljan {\it et al.}~\cite{danelljan2016beyond} propose  continuous convolution filters for correlation filter tracking, and later optimize this method in~\cite{danelljan2017eco}.

\vspace{0.12em}
\noindent{\bf Siamese tracking.} Siamese network has attracted increasing interest for visual tracking because of its balanced accuracy and accuracy. Tao {\it et al.}~\cite{tao2016siamese} utilize Siamese network to off-line learn a matching function from a large set of sequences, then use the fixed matching function to search for the target in a local region. Bertinetto {\it et al.}~\cite{bertinetto2016fully} introduce a fully convolutional Siamese network (SiamFC) for tracking by measuring the region-wise feature similarity between the target object and the candidate. Owing to its light structure and without model update, SiamFC runs efficiently at 80 fps. Held {\it et al.}~\cite{held2016learning} propose the GOTURN approach by learning a motion prediction model with the Siamese network. Valmadre {\it et al.}~\cite{valmadre2017end} use a Siamese network to learn the feature representation for correlation filter tracking. He {\it et al.}~\cite{he2018twofold} introduce a two-fold Siamese network for tracking. Wang {\it et al.}~\cite{wang2018learning} incorporate attention mechanism into Siamese network to learn a more discriminative metric for tracking. Notably, Li {\it et al.}~\cite{li2018high} combine Siamese network with RPN, and propose a one-stage Siamese-RPN tracker, achieving excellent performance. Zhu {\it et al.}~\cite{zhu2018distractor} introduce more negative samples to train a distractor-aware Siamese-RPN tracker. Despite improvement, this approach requires large extra training data from other domains.

\vspace{0.12em}
\noindent{\bf Multi-level features.} The features from different layers in the neural network contain different information. The high-level feature consists of more abstract semantic cues, while the low-level layers contains more detailed spatial information~\cite{long2015fully}. It has been proven that tracking can be benefited using multi-level features. In~\cite{ma2015hierarchical}, Ma {\it et al.} separately use features in three different layers for three correlation models, and fuse their outputs for the final tracking result. Wang {\it et al.}~\cite{wang2015visual} develop two regression models with features from two layers to distinguish similar semantic distractors.

\vspace{0.12em}
\noindent{\bf Our approach.} In this paper, we focus on solving the problem of class imbalance to improve model discriminability. Our approach is related but different from the Siamese-RPN tracker~\cite{li2018high}, which applies one-stage RPN for classification and localization and skips the data imbalance problem. In contrast, our approach cascades a sequence of RPNs to address the data imbalance by performing hard negative sampling, and progressively refines anchor boxes for better target localization using multi-regression. Our method is also related to~\cite{ma2015hierarchical,wang2015visual} using multi-level features for tracking. However, unlike~\cite{ma2015hierarchical,wang2015visual} in which multi-level features are separately used for independent models, we propose a feature transfer block to fuse features across layer for each RPN, improving its discriminative power in distinguishing the target object from complex background.

\section{Siamese Cascaded RPN (C-RPN)}

In this section, we detail the Siamese Cascaded RPN (referred to as C-RPN) as shown in Fig.~\ref{fig:cpn}.

C-RPN contains two subnetworks: the Siamese network and the cascaded RPN. The Siamese network is utilized to extract the features of the target template $\mathbf{x}$ and the search region $\mathbf{z}$. Afterwards, C-RPN receives the features of $\mathbf{x}$ and $\mathbf{z}$ for each RPN. Instead of only using the features from one layer, we apply feature transfer block (FTB) to fuse the features from high-level layers for RPN. An RPN simultaneously performs classification and localization on the feature maps of $\mathbf{z}$. According to the classification scores and regression offsets, we filter out the easy negative anchors (\eg, an anchor whose negative confidence is larger than a preset threshold $\theta$), and refine the locations and sizes of the rest anchors, which are used for training RPN in the next stage.

\subsection{Siamese Network}

\begin{figure}
	\centering
	\includegraphics[width=1\linewidth]{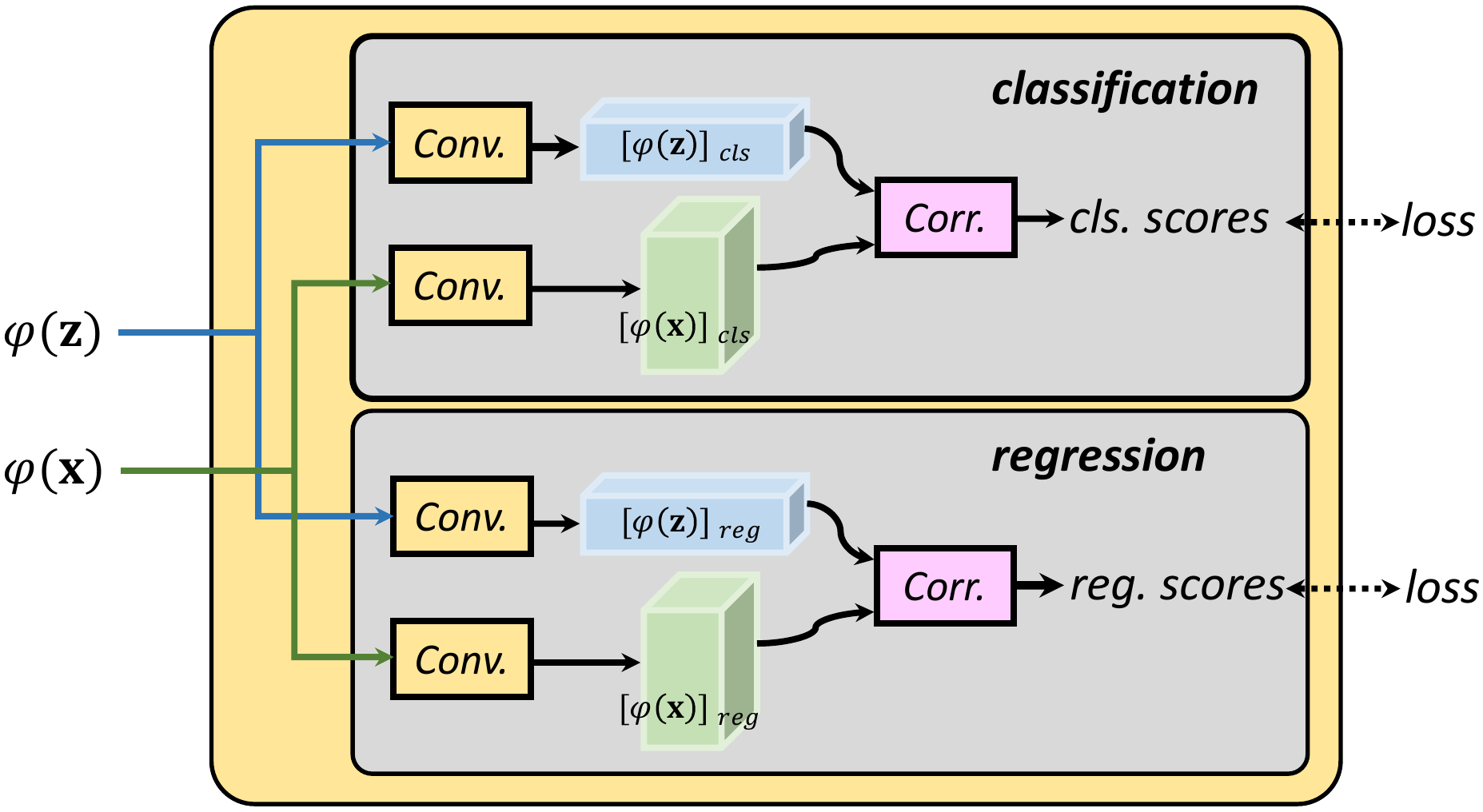}
	\caption{Architecture of RPN. Best viewed in color.}
	\label{fig:rpn}
\vspace{-2mm}
\end{figure}

As in~\cite{bertinetto2016fully}, we adopt the modified AlexNet~\cite{krizhevsky2012imagenet} to develop our Siamese network. The Siamese network comprises two identical branches, the z-branch and the x-branch, which are employed to extract features from the target template $\mathbf{z}$ and the search region $\mathbf{x}$, respectively (see Fig.~\ref{fig:cpn}). The two branches are designed to share parameters to ensure the same transformation applied to both $\mathbf{z}$ and $\mathbf{x}$, which is crucial for the similarity metric learning. More details about the Siamese network can be referred to~\cite{bertinetto2016fully}.

Different from~\cite{li2018high} that only uses the features from the last layer of the Siamese network for tracking, we leverage the features from multiple levels to improve model robustness. For convenience in next, we denote $\varphi_{i}(\mathbf{z})$ and $\varphi_{i}(\mathbf{x})$ as the feature transformations of $\mathbf{z}$ and $\mathbf{x}$ from the conv-$i$ layer in the Siamese network with $N$ layers\footnote{For notation simplicity, we name each layer in the Siamese network in an {\bf inverse} order, \ie, conv-$N$, conv-$(N-1)$, $\cdots$, conv-$2$, conv-$1$ for the low-level to the high-level layers.}.

\subsection{One-Stage RPN in Siamese Network}

Before describing C-RPN, we first review the one-stage Siamese RPN tracker~\cite{li2018high}, which consists of two branches of classification and regression for anchors, as depicted in Fig.~\ref{fig:rpn}. It takes as inputs the feature transformations $\varphi_{1}(\mathbf{z})$ and $\varphi_{1}(\mathbf{x})$ of $\mathbf{z}$ and $\mathbf{x}$ from the last layer of the Siamese network, and outputs classification scores and regression offsets for anchors. For simplicity, we remove the subscripts in feature transformations in next.

To ensure classification and regression for each anchor, two convolution layers are utilized to adjust the channels of $\varphi(\mathbf{z})$ into suitable forms, denoted as $[\varphi(\mathbf{z})]_{cls}$ and $[\varphi(\mathbf{z})]_{reg}$, for classification and regression, respectively. Likewise, we apply two convolution layers for $\varphi(\mathbf{x})$ but keep the channels unchanged, and obtain $[\varphi(\mathbf{x})]_{cls}$ and $[\varphi(\mathbf{x})]_{reg}$. Therefore, the classification scores $\{c_i\}$ and the regression offsets $\{r_i\}$ for each anchor can be computed as
\begin{equation}
\label{eq1}
\begin{split}
\{c_i\} &= \mathrm{corr}([\varphi(\mathbf{z})]_{cls}, [\varphi(\mathbf{x})]_{cls})\\
\{r_i\} &= \mathrm{corr}([\varphi(\mathbf{z})]_{reg}, [\varphi(\mathbf{x})]_{reg})
\end{split}
\end{equation}
where $i$ is the anchor index, and $\rm corr(\mathbf{a}, \mathbf{b})$ denotes correlation between $\mathbf{a}$ and $\mathbf{b}$ where $\mathbf{a}$ is served as the kernel. Each $c_i$ is a 2d vector, representing for negative and positive confidences of the $i^{\rm th}$ anchor. Similarly, each $r_i$ is a 4d vector which represents the offsets of center point location and size of the anchor to groundtruth. Siamese RPN is trained with a multi-task loss consisting of two parts, \ie, the classification loss (\ie, softmax loss) and the regression loss (\ie, smooth $L_1$ loss). We refer readers to~\cite{li2018high,ren2015faster} for further details.

\subsection{Cascaded RPN}

As mentioned earlier, previous Siamese trackers mostly ignore the problem of class imbalance, resulting in degenerated performance in presence of similar semantic distractors. Besides, they only use the high-level semantic features from the last layer, which does not fully explore multi-level features. To address these issues, we propose a multi-stage tracking framework by cascading a set of $L$ ($L \le N$) RPNs.

For RPN$_l$ in the $l^{\rm th}$ ($1 < l \le L$) stage, it receives fused features $\Phi_l(\mathbf{z})$ and $\Phi_l(\mathbf{x})$ of the conv-$l$ layer and the high-level layers from FTB, instead of features $\varphi_l(\mathbf{z})$ and $\varphi_l(\mathbf{x})$ from a single separate layer~\cite{li2018high,bertinetto2016fully}. The $\Phi_l(\mathbf{z})$ and $\Phi_l(\mathbf{x})$ are obtained as follows,
\begin{equation}
\label{eq4}
\begin{split}
\Phi_{l}(\mathbf{z}) &= \mathrm{FTB} \big(\Phi_{l-1}(\mathbf{z}), \varphi_{l}(\mathbf{z})\big) \\
\Phi_{l}(\mathbf{x}) &= \mathrm{FTB} \big(\Phi_{l-1}(\mathbf{x}), \varphi_{l}(\mathbf{x})\big)
\end{split}
\end{equation}
where $\mathrm{FTB}(\cdot,\cdot)$ denotes the FTB as described in Section~\ref{ftb}. For RPN$_1$, $\Phi_{1}(\mathbf{z}) = \varphi_{1}(\mathbf{z})$ and $\Phi_{1}(\mathbf{x}) = \varphi_{1}(\mathbf{x})$. Therefore, the classification scores $\{c_i^{l}\}$ and the regression offsets $\{r_i^{l}\}$ for anchors in stage $l$ are calculated as
\begin{equation}
\label{eq5}
\begin{split}
\{c_i^{l}\} &= \mathrm{corr}([\Phi_l(\mathbf{z})]_{cls}, [\Phi_l(\mathbf{x})]_{cls})\\
\{r_i^{l}\} &= \mathrm{corr}([\Phi_l(\mathbf{z})]_{reg}, [\Phi_l(\mathbf{x})]_{reg})
\end{split}
\end{equation}
where $[\Phi_{l}(\mathbf{z})]_{cls}$, $[\Phi_{l}(\mathbf{x})]_{cls}$, $[\Phi_{l}(\mathbf{z})]_{reg}$ and $[\Phi_{l}(\mathbf{x})]_{reg}$ are derived by performing convolutions on $\Phi_{l}(\mathbf{z})$ and $\Phi_{l}(\mathbf{x})$.

\begin{figure}
	\centering
	\includegraphics[width=0.47\linewidth]{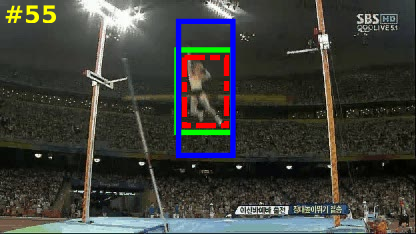} \includegraphics[width=0.47\linewidth]{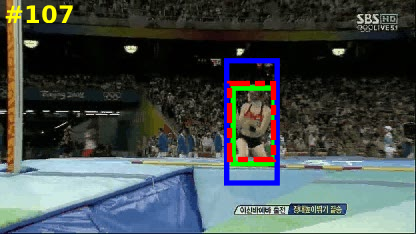} \\
	\includegraphics[width=0.8\linewidth]{fig1_mark.pdf} \\
	\caption{Localization using a single regressor and multiple regressors.The multiple regressors in C-RPN can better handle large scale changes for more accurate localization. Best viewed in color.}
	\label{fig:4}
	\vspace{-2mm}
\end{figure}

Let $A_{l}$ denote the anchor set in stage $l$. With classification scores $\{c_i^{l}\}$, we can filter out anchors in $A_{l}$ whose negative confidences are larger than a preset threshold $\theta$, and the rest are formed into a new set of anchor $A_{l+1}$, which is employed for training RPN$_{l+1}$. For RPN$_1$, $A_1$ is pre-defined. Besides, in order to provide a better initialization for regressor of RPN$_{l+1}$, we refine the center locations and sizes of anchors in $A_{l+1}$ using the regression results $\{r_i^{l}\}$ in RPN$_{l}$, thus generate more accurate localization compared to a single step regression in Siamese RPN~\cite{li2018high}, as illustrated in Fig.~\ref{fig:4}. Fig.~\ref{fig:cpn} shows the cascade architecture of C-RPN.

\begin{figure}
	\centering
	\includegraphics[width=\linewidth]{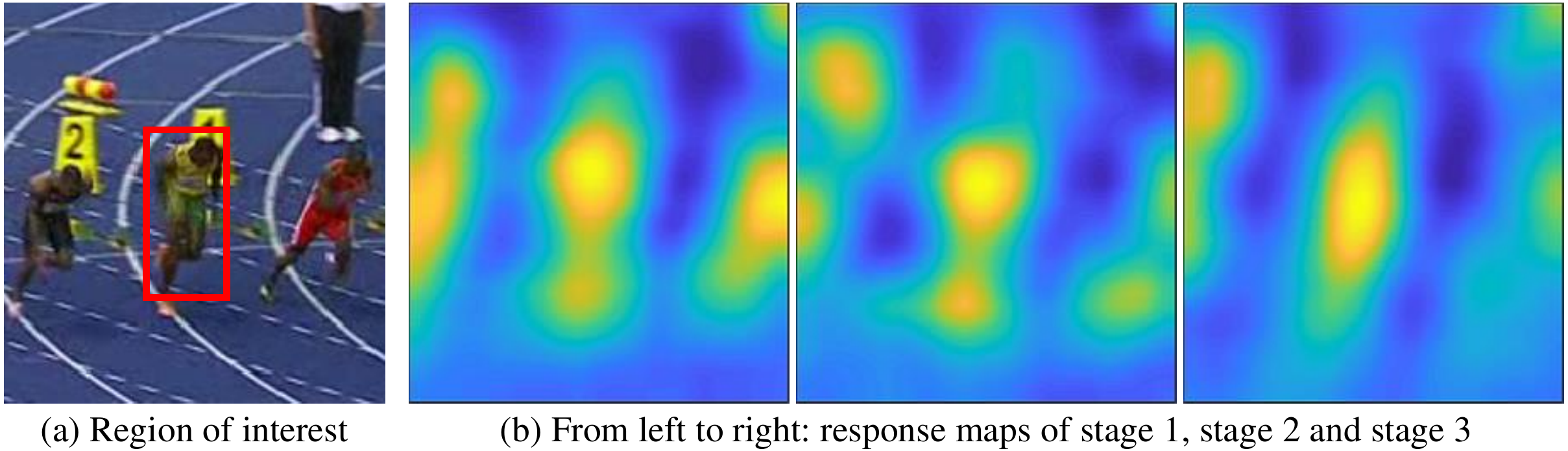}\\
	\caption{Response maps in different stages. Image (a) is the region of interest, and (b) shows the response maps obtained by RPN in three stages. We can see that RPN is sequentially more discriminative in distinguishing distractors. Best viewed in color.}
	\label{fig:5}
	\vspace{-2mm}
\end{figure}

The loss function $\ell_{{\rm RPN}_{l}}$ for RPN$_l$ is composed of classification loss function $L_{\rm cls}$ (softmax loss) and regression loss function $L_{\rm loc}$ (smooth $L_1$ loss) as follows,
\begin{equation}
\label{rpn_loss}
\ell_{{\rm RPN}_{l}}(\{c_i^{l}\}, \{r_i^{l}\}) = \sum_{i}L_{\rm cls}(c_i^{l}, c_i^{l*}) + \lambda \sum_{i}c_{i}^{l*}L_{\rm loc}(r_i^{l}, r_i^{l*})
\end{equation}
where $i$ is the anchor index in $A_l$ of stage $l$, $\lambda$ a weight to balance losses, $c_{i}^{l*}$ the label of anchor $i$, and $r_{i}^{l*}$ the true distance between anchor $i$ and groundtruth. Following~\cite{ren2015faster}, $r_{i}^{l*}=(r_{i(\rm x)}^{l*}, r_{i (\rm y)}^{l*}, r_{i (\rm w)}^{l*}, r_{i (\rm h)}^{l*})$ is a 4d vector, such that
\begin{equation}
\begin{aligned}
r_{i(\rm x)}^{l*} &= (x^*-x_a^{l})/w_a^{l}  & r_{i(\rm y)}^{l*} &= (y^*-y_a^{l})/h_a^{l} \\
r_{i(\rm w)}^{l*} &= {\rm log}(w^*/w_a^{l}) & r_{i(\rm h)}^{l*} &= {\rm log}(y^*/h_a^{l})
\end{aligned}
\end{equation}
where $x$, $y$, $w$ and $h$ are center coordinates of a box and its width and height. Variables $x^*$ and $x_a^{l}$ are for groundtruth and anchor of stage $l$ (likewise for $y$, $w$ and $h$). It is worth noting that, different from~\cite{li2018high} using fixed anchors, the anchors in C-RPN are progressively adjusted by the regressor in the previous stage, and computed as
\begin{equation}
\label{eq8}
\begin{aligned}
x_a^{l} &=x_a^{l} + w_{a}^{l-1}   r_{i(\rm x)}^{l-1}    & y_a^{l}  &=y_a^{l} + h_{a}^{l-1}    r_{i(\rm y)}^{l-1}  \\
w_a^{l} &= w_a^{l-1}     {\rm exp}(r_{i(\rm w)}^{l-1})  & h_a^{l} &= h_a^{l-1}    {\rm exp}(r_{i(\rm h)}^{l-1})
\end{aligned}
\end{equation}
For the anchor in the first stage, $x_a^{1}$, $y_a^{1}$, $w_a^{1}$ and $h_a^{1}$ are pre-defined.

The above procedure forms the proposed cascaded RPN. Due to the rejection of easy negative anchors, the distribution of training samples for each RPN is gradually more balanced. As a result, the classifier of each RPN is sequentially more discriminative in distinguishing difficult distractors. Besides, multi-level feature fusion further improves the discriminability in handing complex background. Fig.~\ref{fig:5} shows the discriminative powers of different RPNs by demonstrating detection response map in each stage.

The loss function $\ell_{\rm CRPN}$ of C-RPN consists of the loss functions of all RPN$_l$. For each RPN, loss function is computed using Eq. (\ref{rpn_loss}), and $\ell_{\rm CRPN}$ is expresses as
\begin{equation}
\label{crpn_loss}
\ell_{\rm CRPN} = \sum_{l=1}^{L} \ell_{\rm RPN_{l}}
\end{equation}

\subsection{Feature Transfer Block}
\label{ftb}

\begin{figure}[t]
	\centering
	\includegraphics[width=\linewidth]{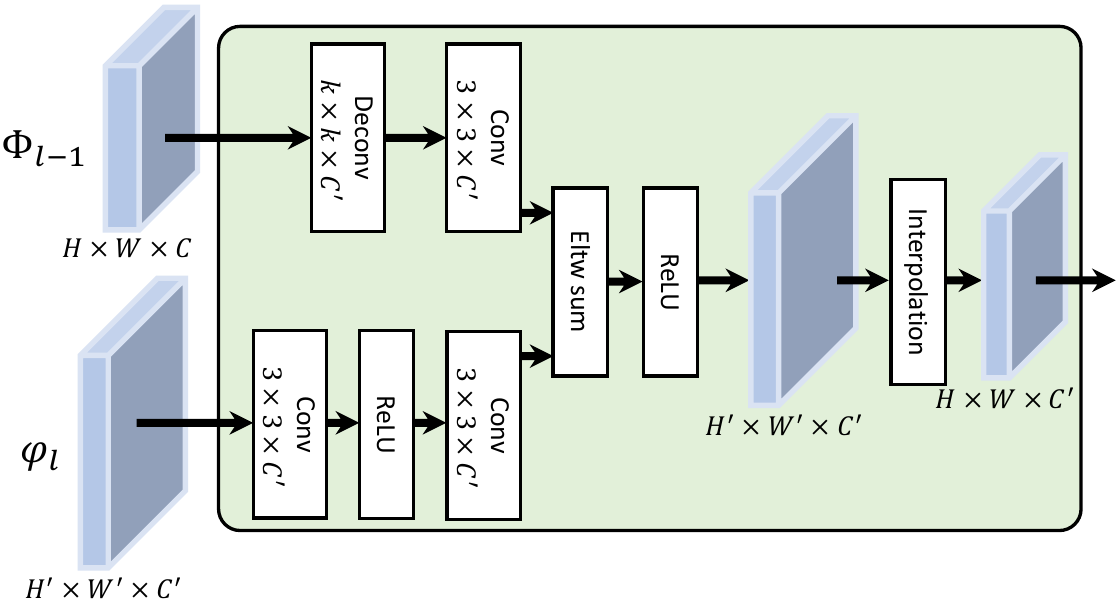} \\
	\caption{Overview of feature transfer block. Best viewed in color.}
	\label{fig:6}
\vspace{-2mm}
\end{figure}

To effectively leverage multi-level features, we introduce FTB to fuse features across layers so that each RPN is able to share high-level semantic feature to improve the discriminability. In detail, a deconvolution layer is used to match the feature dimensions of different sources. Then, different features are fused using element-wise summation, followed a ReLU layer. In order to ensure the same groundtruth for anchors in each RPN, we apply the interpolation to rescale the fused features such that the output classification maps and regression maps have the same resolution for all RPN. Fig.~\ref{fig:6} shows the feature transferring for RPN$_{l}$ ($l>1$).

\subsection{Training and Tracking}
\noindent
{\bf Training.} The training of C-RPN is performed on the image pairs that are sampled within a random interval from the same sequence as in~\cite{li2018high}. The multi-task loss function in Eq. (\ref{crpn_loss}) enables us to train C-RPN in an end-to-end manner. Considering that the scale of target changes smoothly in two consecutive frames, we employ one scale with different ratios for each anchor. The ratios of anchors are set to $[0.33, 0.5, 1, 2, 3]$ as in~\cite{li2018high}.

For each RPN, we adopt the strategy as in object detection~\cite{ren2015faster} to determine positive and negative training samples. We define the positive samples as anchors whose Intersection over union (IOU) with groundtruth is larger than a threshold $\tau_{\rm pos}$, and negative samples as anchors whose IoU with groundtruth bounding box is less than a threshold $\tau_{\rm neg}$. We generate at most 64 samples from one image pair.

\begin{algorithm}[!t]\small
	\caption{Tracking with C-RPN}\label{crpn_alg}
	{\bf Input:} frame sequences $\{\mathbf{X}_{t}\}_{t=1}^{T}$ and groundtruth bounding box $\mathbf{b}_1$ of $\mathbf{X}_1$, trained model C-RPN\;
	{\bf Output:} Tracking results $\{\mathbf{b}_{t}\}_{t=2}^{T}$\;
	Extract target template $\mathbf{z}$ in $\mathbf{X}_1$ using  $\mathbf{b}_1$ \;
	Extract features $\{\varphi_{l}(\mathbf{z})\}_{l=1}^{L}$ for $\mathbf{z}$ from C-RPN\;
	Initialize anchors $A_1$\;
	\For{$t=2$ {\rm to} $T$}
	{
		Extract the search region $\mathbf{x}$ in $\mathbf{X}_t$ using  $\mathbf{b}_{t-1}$ \;
		Extract features $\{\varphi_{l}(\mathbf{x})\}_{l=1}^{L}$ for $\mathbf{x}$ from C-RPN\;
		\For {$l=1$ {\rm to} $L$}
		{
			\eIf {$l$ {\rm equals to 1}}
			{
				$\Phi_{l}(\mathbf{z}) = \varphi_{l}(\mathbf{z})$, $\Phi_{l}(\mathbf{x}) = \varphi_{l}(\mathbf{x})$\;
			}
			{
				$\Phi_{l}(\mathbf{z})$, $\Phi_{l}(\mathbf{x})$ $\leftarrow$ Eq. (\ref{eq4}) \;
			}
			$\{c_{i}^{l}\}$, $\{r_{i}^{l}\}$ $\leftarrow$ Eq. (\ref{eq5})\;
			Remove any anchor $i$ from $A_l$ whose negative confidence $c_{i(\rm neg)}^{l} > \theta$ \;
			$A_{l+1}$ $\leftarrow$ Refine the rest anchors in $A_{l}$ with $\{r_{i}^{l}\}$ using Eq. (\ref{eq8})\;
		}
		Target proposals $\leftarrow$ $A_{L+1}$ \;
		Selet the best proposal as tracking result $\mathbf{b}_{k}$ using strategies in~\cite{li2018high}\;
	}
\end{algorithm}

\vspace{0.1 em}
\noindent {\bf Tracking.} We formulate tracking as multi-stage detection. For each video, we pre-compute feature embeddings for the target template in the first frame. In a new frame, we extract a region of interest according to the result in last frame, and then perform detection using C-RPN on this region. In each stage, an RPN outputs the classification scores and regression offsets for anchors. The anchors with negative scores lager then $\theta$ are discarded, and the rest are refined and taken over by RPN in next stage. After the last stage $L$, the remained anchors are regarded as target proposals, from which we determine the best one as the final tracking result using strategies in~\cite{li2018high}. Alg.~\ref{crpn_alg} summarizes the tracking process by C-RPN.

\section{Experiments}
\noindent
{\bf Implementation detail.} C-RPN is implemented in Matlab using MatConvNet~\cite{vedaldi2015matconvnet} on a single Nvidia GTX 1080 with 8GB memory. The backbone Siamese network adopts the modified AlexNet~\cite{krizhevsky2012imagenet} by removing group convolutions. Instead of training from scratch, we borrow the parameters from the pretrained model on ImageNet~\cite{deng2009imagenet}. During training, the parameters of first two layers are frozen. The number $L$ of stages is set to 3. The thresholds $\theta$, $\tau_{\rm pos}$ and $\tau_{\rm neg}$ are empirically set to 0.95, 0.6 and 0.3. C-RPN is trained end-to-end over 50 epochs using SGD, and the learning rate is annealed geometrically at each epoch from $10^{-2}$ to $10^{-6}$. We train C-RPN using the training data from~\cite{fan2018lasot} for experiment under Protocol \uppercase\expandafter{\romannumeral2} on LaSOT~\cite{fan2018lasot}, and using VID~\cite{russakovsky2015imagenet} and YT-BB~\cite{real2017youtube} for other experiments.

Note that the comparison with Siamese-RPN~\cite{li2018high} is fair since the same training data is used for training.

\subsection{Experiments on OTB-2013 and OTB-2015}

\begin{figure}[t]
	\centering
	\includegraphics[width=0.495\linewidth]{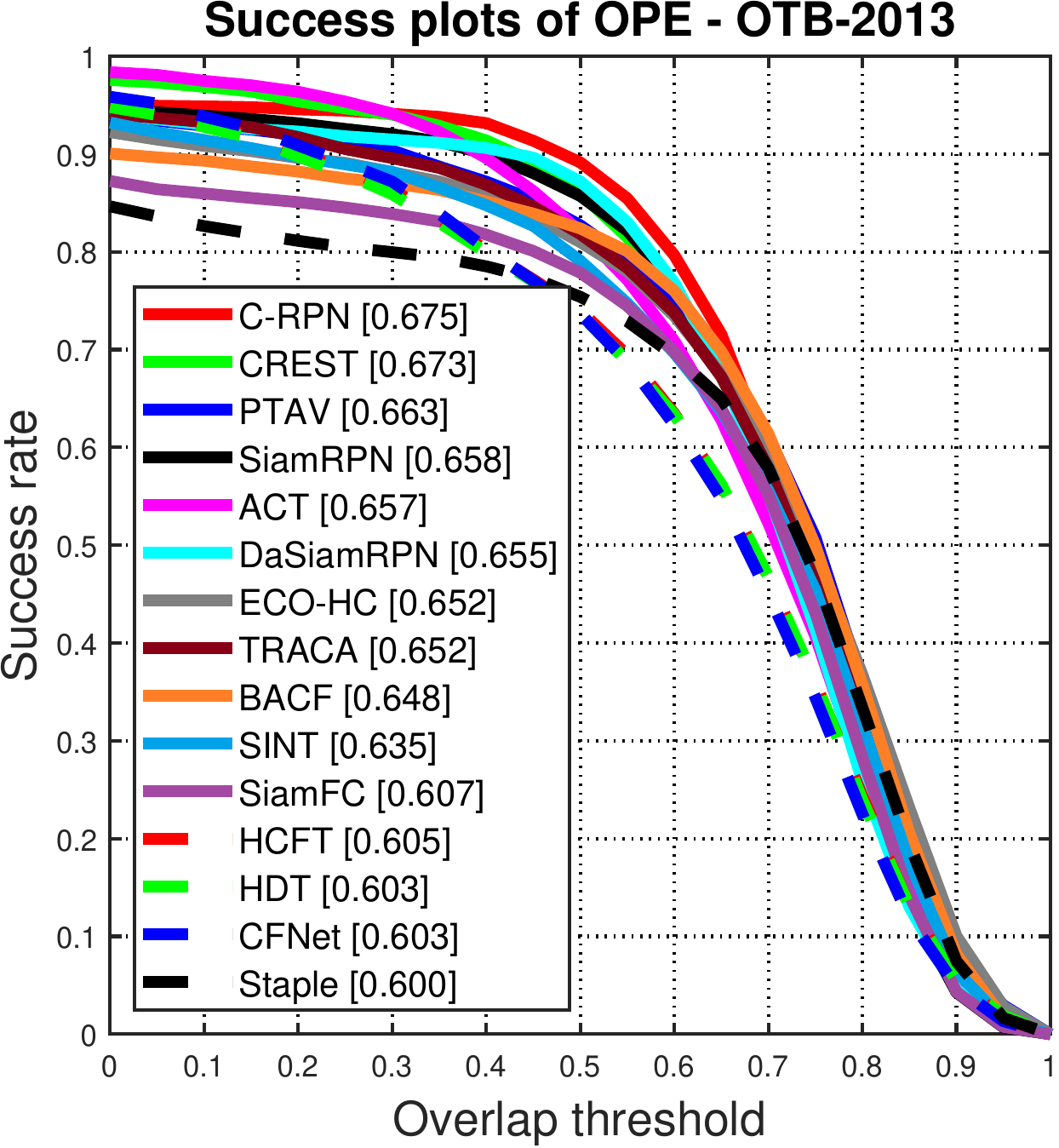} \includegraphics[width=0.495\linewidth]{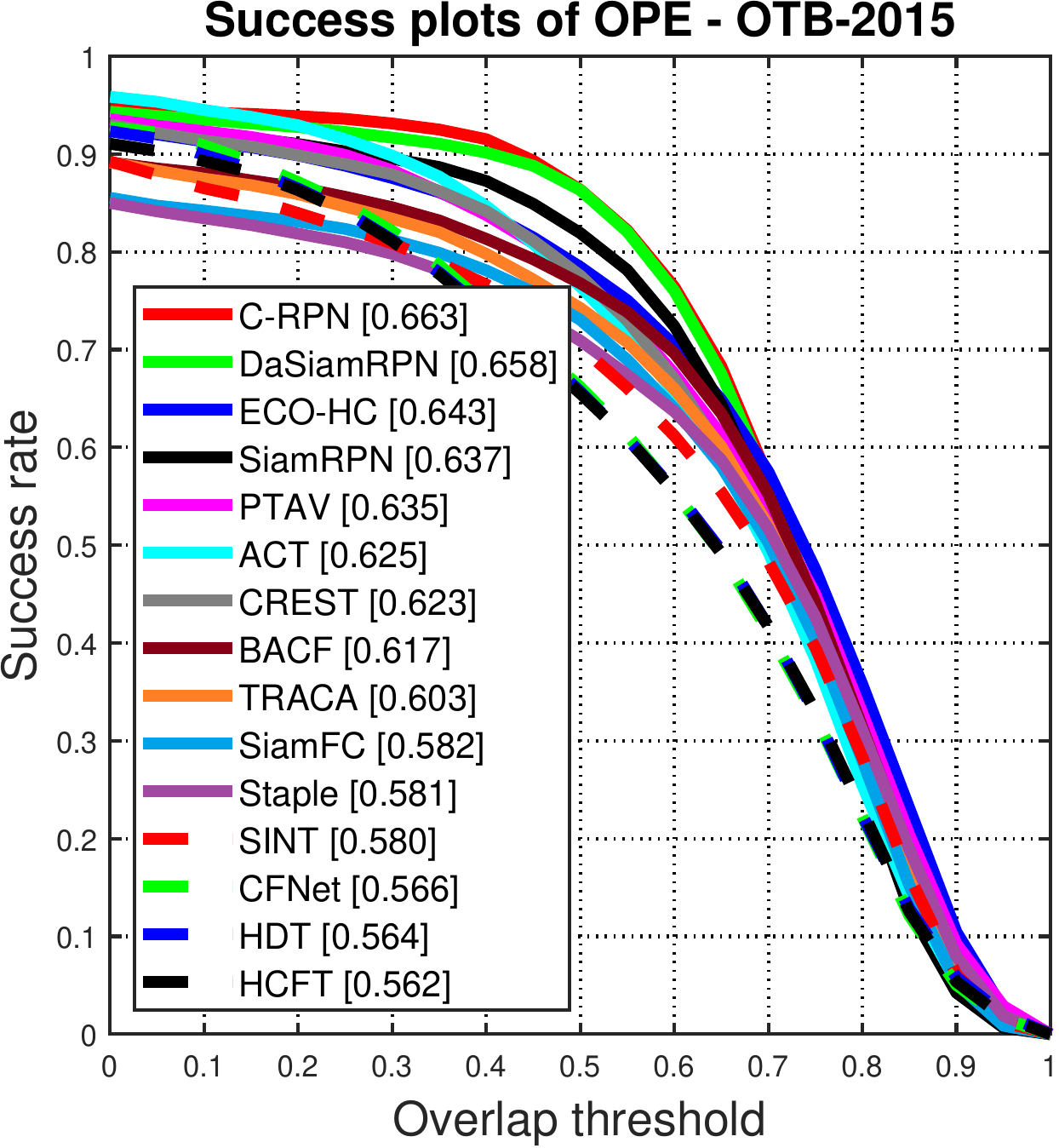}
	\caption{Comparisons with stage-of-the-art tracking approaches on OTB-2013~\cite{wu2013online} and OTB-2015~\cite{wu2015object}. C-RPN achieves the best results on both benchmarks. Best viewed in color.}
	\label{fig:otb}
\vspace{-2mm}
\end{figure}

We conduct experiments on the popular OTB-2013~\cite{wu2013online} and OTB-2015~\cite{wu2015object} which consist of 51 and 100 fully annotated videos, respectively. C-RPN runs at around 36 fps.

Following~\cite{wu2013online}, we adopt the {\it precision} plot in {\it one-pass evaluation} (OPE) to assess different trackers. The comparison with 14 state-of-the-art trackers (SiamRPN~\cite{li2018high}, DaSiamRPN~\cite{zhu2018distractor}, TRACA~\cite{choi2018context}, ACT~\cite{chen2018real}, BACF~\cite{galoogahi2017learning}, ECO-HC~\cite{danelljan2017eco}, CREST~\cite{song2017crest},  SiamFC~\cite{bertinetto2016fully}, Staple~\cite{bertinetto2016staple}, PTAV~\cite{fan2017parallel}, SINT~\cite{tao2016siamese}, CFNet~\cite{valmadre2017end}, HDT~\cite{qi2016hedged} and HCFT~\cite{ma2015hierarchical}) is shown in Fig.~\ref{fig:otb}. C-RPN achieves the best performance on both two benchmarks. In specific, we obtain the 0.675 and 0.663 precision scores on OTB-2013 and OTB-2015, respectively. In comparison with the baseline one-stage SiamRPN with 0.658 and 0.637 precision scores, we obtain improvements by 1.9\% and 2.6\%, showing the advantages of multi-stage RPN in accurate localization. DaSiamRPN uses extra negative training data from other domains to improve the ability to handle similar distractors, and obtains 0.655 and 0.658 precision scores. Without using extra training data, C-RPN outperforms DaSiamRPN by 2.0\% and 0.5\%. More results and comparisons on OTB-2013~\cite{wu2013online} and OTB-2015~\cite{wu2015object} are shown in the supplementary material.

\subsection{Experiments on VOT-2016 and VOT-2017}

\textbf{VOT-2016}~\cite{kristan2016visual} consists of 60 sequences, aiming at assessing the short-term performance of trackers. The overall performance of a tracking algorithm is evaluated using Expected Average Overlap (EAO) which takes both accuracy and robustness into account. The speed of a tracker is represented with a normalized speed (EFO).

\begin{figure}[t]
	\centering
	\includegraphics[width=\linewidth]{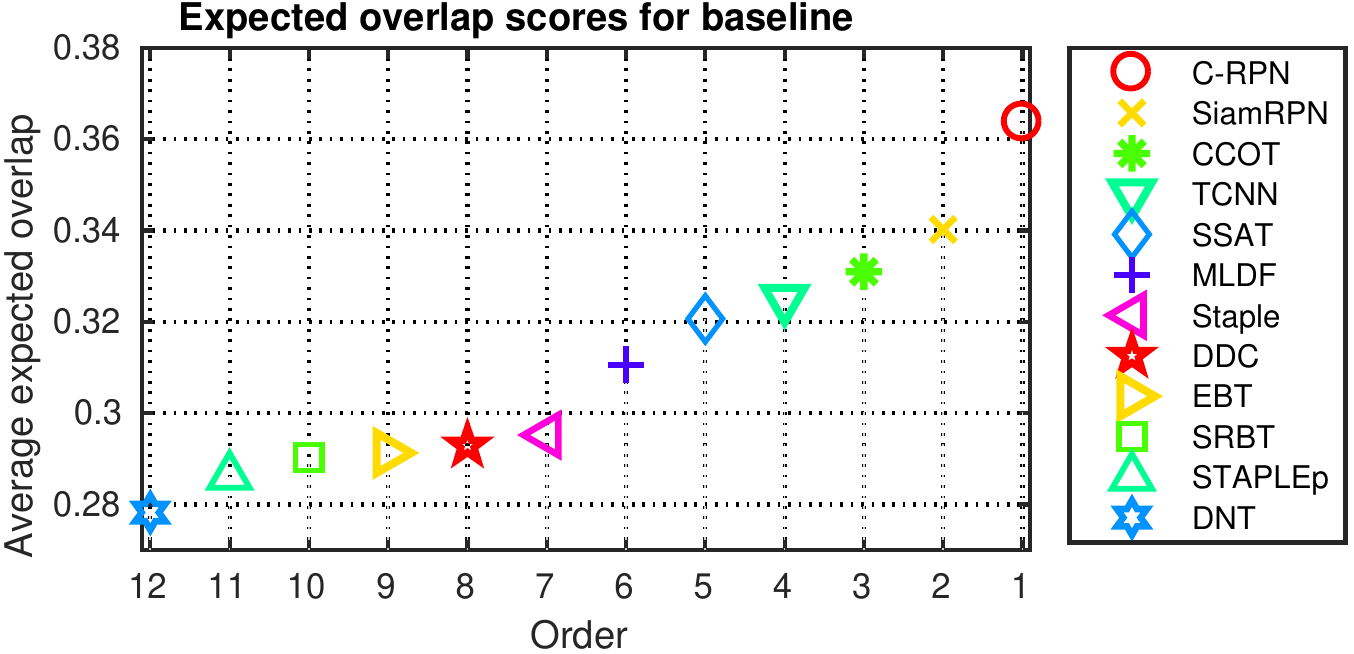}
	\caption{Comparisons on VOT-2016~\cite{kristan2016visual}. Larger (right) value indicates better performance. Our C-RPN significantly outperforms the baseline and other approaches. Best viewed in color.}
	\label{fig:vot16}
\end{figure}

We evaluate C-RPN on VOT-2016, and compare it with 11 trackers including the baseline SiamRPN~\cite{li2018high} and other top ten approaches in VOT-2016. Fig.~\ref{fig:vot16} shows the EAO of different trackers. C-RPN achieves the best results, significantly outperforming the baseline SiamRPN and other approaches. Tab.~\ref{tab:vot16} lists the detailed comparisons of different trackers on VOT-2016. From Tab.~\ref{tab:vot16}, we can see that C-RPN outperforms other trackers in both accuracy and robustness, and runs efficiently.

\renewcommand\arraystretch{1.05}
\begin{table}[t]\small
	\centering
	\caption{Detailed comparisons on VOT-2016~\cite{kristan2016visual}. The best two results are highlighted in \textcolor{red}{\bf red} and \textcolor{blue}{\bf blue} fonts, respectively.}
	\begin{tabular}{rcccc}
		\hline
		Tracker & EAO   & Accuracy & Failure & EFO \\
		\hline \hline
		C-RPN & \textcolor[rgb]{ 1,  0,  0}{\textbf{0.363}} & \textcolor[rgb]{ 1,  0,  0}{\textbf{0.594}} & 0.95  & 9.3 \\
		\hline
		SiamRPN~\cite{li2018high} & \textcolor[rgb]{ 0,  0,  1}{\textbf{0.344}} & 0.560 & 1.12  & \textcolor[rgb]{ 0,  0,  1}{\textbf{23.0}} \\
		C-COT~\cite{danelljan2016beyond} & 0.331 & 0.539 & \textcolor[rgb]{ 0,  0,  1}{\textbf{0.85}} & 0.5 \\
		TCNN~\cite{kristan2016visual}  & 0.325 & 0.554 & 0.96  & 1.1 \\
		SSAT~\cite{kristan2016visual}  & 0.321 & \textcolor[rgb]{ 0,  0,  1}{\textbf{0.577}} & 1.04  & 0.5 \\
		MLDF~\cite{kristan2016visual}  & 0.311 & 0.490 & \textcolor[rgb]{ 1,  0,  0}{\textbf{0.83}} & 1.2 \\
		Staple~\cite{bertinetto2016staple} & 0.295 & 0.544 & 1.35  & 11.1 \\
		DDC~\cite{kristan2016visual}   & 0.293 & 0.541 & 1.23  & 0.2 \\
		EBT~\cite{zhu2016beyond}   & 0.291 & 0.465 & 0.90  & 3.0 \\
		SRBT~\cite{kristan2016visual}  & 0.290 & 0.496 & 1.25  & 3.7 \\
		STAPLEp~\cite{kristan2016visual} & 0.286 & 0.557 & 1.32  & \textcolor[rgb]{ 1,  0,  0}{\textbf{44.8}} \\
		DNT~\cite{chi2017dual}   & 0.278 & 0.515 & 1.18  & 1.1 \\
		\hline
	\end{tabular}%
	\label{tab:vot16}%
\vspace{-2mm}
\end{table}%

\textbf{VOT-2017}~\cite{kristan2017visual} contains 60 sequences, which are developed by replacing the least 10 challenging videos in VOT-2016~\cite{kristan2016visual} with 10 difficult sequences. Different from VOT-2016~\cite{kristan2016visual}, VOT-2017~\cite{kristan2017visual} introduces a new real-time experiment by taking into both tracking performance and efficiency. We compare C-RPN with SiamRPN~\cite{li2018high} and other top ten approaches in VOT-2017 using the EAO of baseline and real-time experiments, as shown in Tab.~\ref{tab:vot17}. From Tab.~\ref{tab:vot17}, C-RPN achieves a EAO score of 0.289, which significantly outperforms the one-stage SiamRPN~\cite{li2018high} with EAO score of 0.243. In addition, compared with LSART~\cite{sun2018learning} and CFWCR~\cite{kristan2017visual}, C-RPN shows competitive performance. In real-time experiment, C-RPN obtains the best result with EAO score of 0.273, outperforming all other trackers.

\renewcommand\arraystretch{1.05}
\begin{table}[t]\small
	\centering
	\caption{Comparisons on VOT-2017~\cite{kristan2017visual}. The best two results are highlighted in \textcolor{red}{\bf red} and \textcolor{blue}{\bf blue} fonts, respectively.}
	\begin{tabular}{rcc}
		\hline
		Tracker & \tabincell{c}{Baseline EAO}  & \tabincell{c}{Real-time EAO} \\
		\hline\hline
		C-RPN & 0.289 & \textcolor[rgb]{ 1,  0,  0}{\textbf{0.273}} \\
		\hline
		SiamRPN~\cite{li2018high} & 0.243 & \textcolor[rgb]{ 0,  0,  1}{\textbf{0.244}} \\
		LSART~\cite{sun2018learning} & \textcolor[rgb]{ 1,  0,  0}{\textbf{0.323}} & 0.055 \\
		CFWCR~\cite{kristan2017visual} & \textcolor[rgb]{ 0,  0,  1}{\textbf{0.303}} & 0.062 \\
		CFCF~\cite{gundogdu2018good}  & 0.286 & 0.059 \\
		ECO~\cite{danelljan2017eco}   & 0.280 & 0.078 \\
		Gnet~\cite{kristan2017visual}  & 0.274 & 0.060 \\
		MCCT~\cite{kristan2017visual}  & 0.270 & 0.061 \\
		C-COT~\cite{danelljan2016beyond} & 0.267 & 0.058 \\
		CSRDCF~\cite{lukezic2017discriminative} & 0.256 & 0.100 \\
		SiamDCF~\cite{kristan2017visual} & 0.249 & 0.135 \\
		MCPF~\cite{zhang2017multi}  & 0.248 & 0.060 \\
		\hline
	\end{tabular}%
	\label{tab:vot17}%
\vspace{-2mm}
\end{table}%

\subsection{Experiment on LaSOT}
LaSOT~\cite{fan2018lasot} is a recent large-scale dataset aiming at both training and evaluating trackers. We compare C-RPN to 35 approaches, including ECO~\cite{danelljan2017eco}, MDNet~\cite{nam2016learning}, SiamFC~\cite{bertinetto2016fully}, VITAL~\cite{song2018vital}, StructSiam~\cite{zhang2018structured}, TRACA~\cite{choi2018context}, BACF~\cite{galoogahi2017learning} and so forth. We refer readers to~\cite{fan2018lasot} for more details about the compared trackers. We do not compare C-RPN to Siamese-RPN~\cite{li2018high} because neither its implementation nor results on LaSOT are available.

\begin{figure*}[!t]
	\centering
	\includegraphics[width=0.49\linewidth]{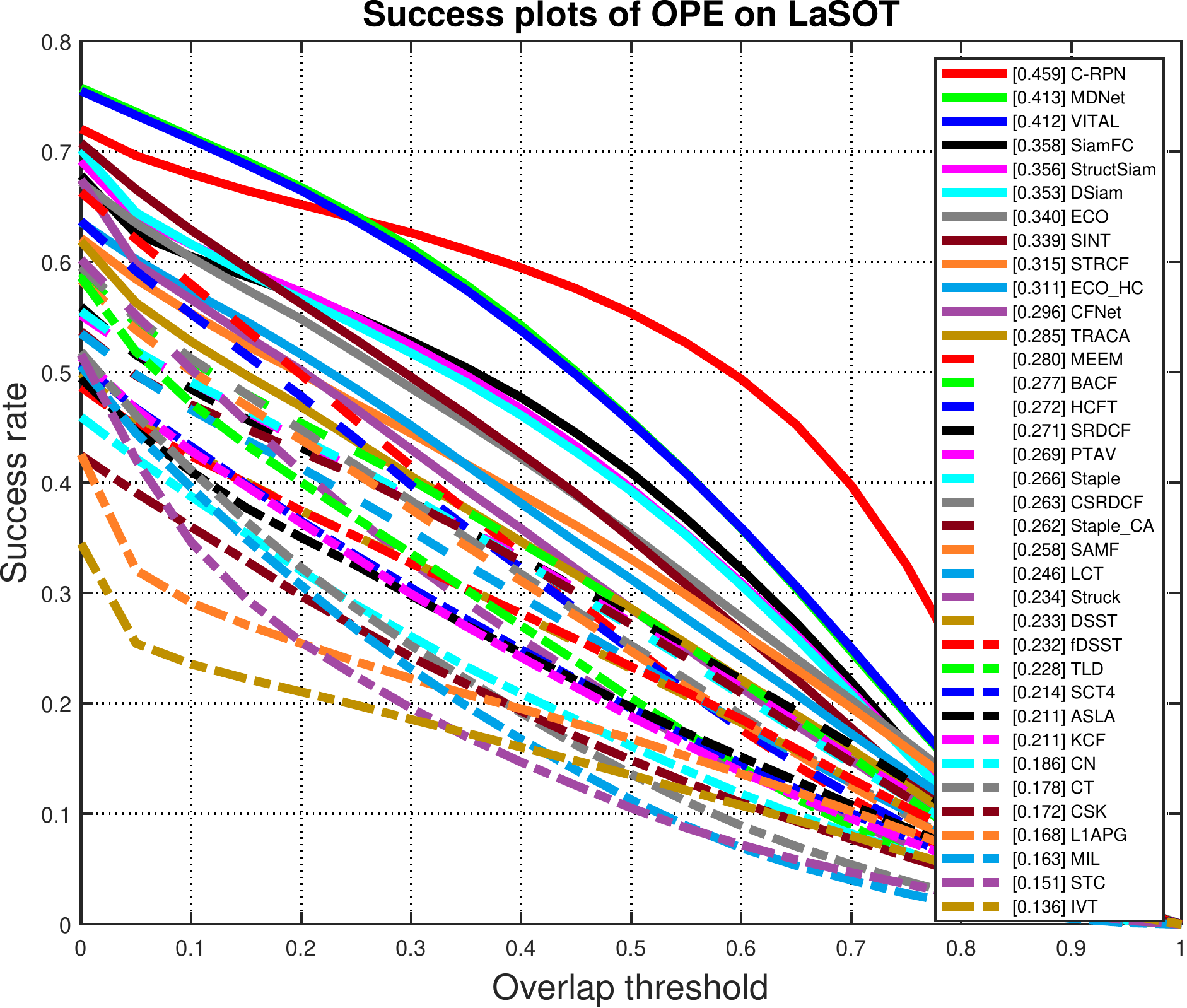}\includegraphics[width=0.49\linewidth]{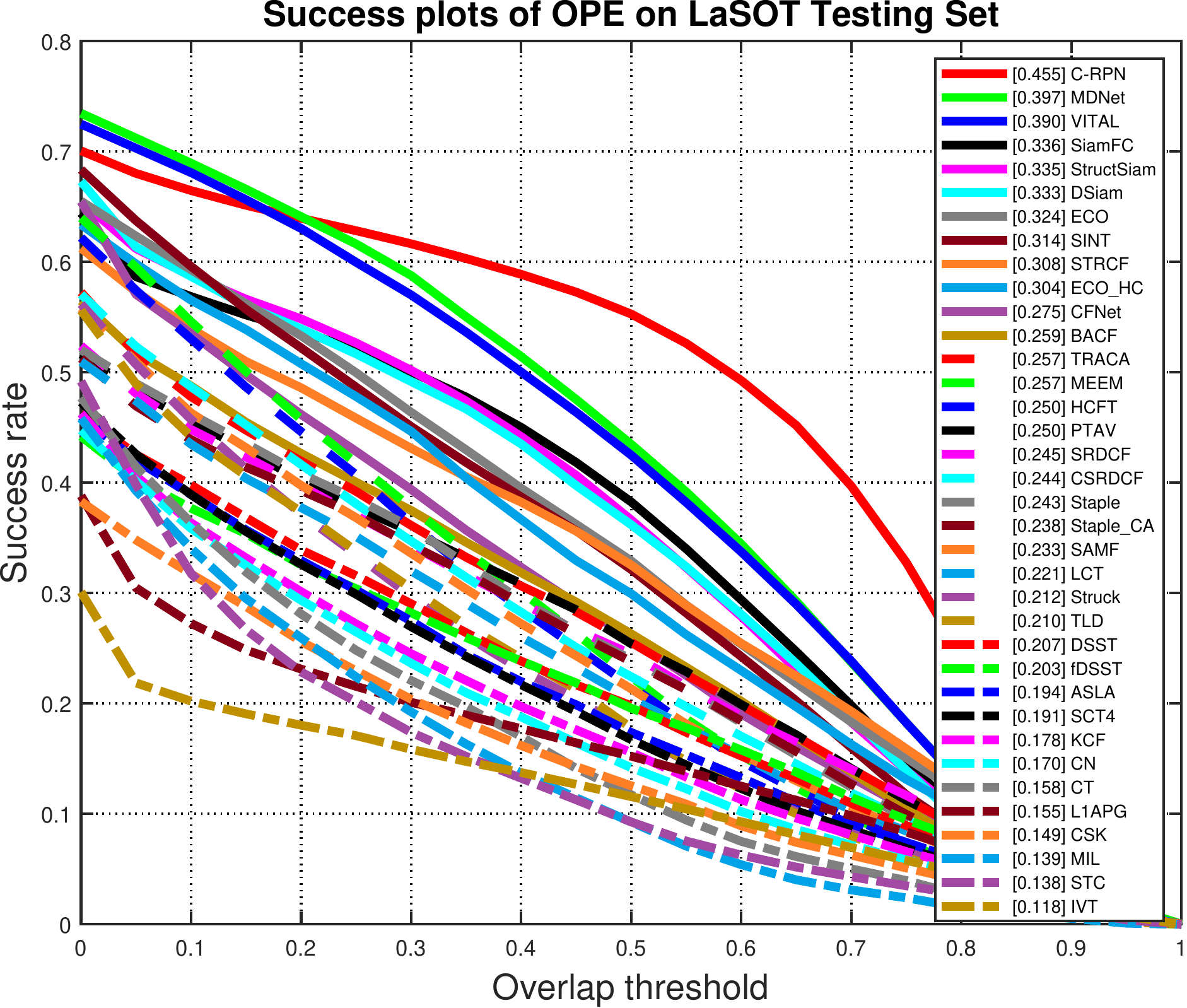}
	\caption{Comparisons with state-of-the-art tracking methods on LaSOT~\cite{fan2018lasot}. C-RPN outperforms existing approaches on success by large margins under all two protocols. Best viewed in color.}
	\label{fig:lasot}
\vspace{-2mm}
\end{figure*}

Following~\cite{fan2018lasot}, we report the results of {\it success} (SUC) for different trackers as shown in Fig.~\ref{fig:lasot}. It shows that our C-RPN outperforms all other state-of-the-art trackers under two protocols. We achieve SUC scores of 0.459 and 0.455 under protocol \uppercase\expandafter{\romannumeral1} and \uppercase\expandafter{\romannumeral2}, outperforming the second best tracker MDNet with SUC scores 0.413 and 0.397 by 4.6\% and 5.8, respectively. In addition, C-RPN runs at around 23 fps on LaSOT, which is more efficient than MDNet with around 1 fps. Compared with the Siamese network-based tracker SiamFC with 0.358 and 0.336 SUC scores, C-RPN gains the improvements by 11.1\% and 11.9\%. Due to limited space, we refer readers to supplementary material for more details about results and comparisons on LaSOT.

\subsection{Experiment on TrackingNet}

TrackingNet~\cite{muller2018trackingnet} is proposed to assess the performance of a tracker in the wild. We evaluate C-RPN on its testing set with 511 videos. Following~\cite{muller2018trackingnet}, we use three metrics {\it precision} (PRE), {\it normalized precision} (NPRE) and {\it success} (SUC) for evaluation. Tab.~\ref{tab:tn} demonstrates the comparison results to trackers with top PRE scores\footnote{The result of C-RPN on TrackingNet~\cite{muller2018trackingnet} is evaluated by the server provided by the organizer at \url{http://eval.tracking-net.org/web/challenges/challenge-page/39/leaderboard/42}. The results of compared trackers are reported from~\cite{muller2018trackingnet}. Full comparison is shown in the supplementary material.}, showing that C-RPN achieves the best results on all three metrics. In specific, C-RPN obtains the PRE score of 0.619, NPRE score of 0.746 and SUC score of 0.669, outperforming the second best tracker MDNet with PRE score of 0.565, NPRE score of 0.705 and SUC score of 0.606 by 5.4\%, 4.1\% and 6.3\%, respectively. Besides, C-RPN runs efficiently at a speed of around 32 fps.

\renewcommand\arraystretch{1.05}
\begin{table}[t]\small
	\centering
	\caption{Comparisons on TrackingNet~\cite{muller2018trackingnet} with the best two results highlighted in \textcolor{red}{\bf red} and \textcolor{blue}{\bf blue} fonts, respectively.}
	\begin{tabular}{rccc}
		\hline
		& PRE   & NPRE  & SUC \\
		\hline \hline
		C-RPN & \textcolor[rgb]{ 1,  0,  0}{\textbf{0.619}}  & \textcolor[rgb]{ 1,  0,  0}{\textbf{0.746}} & \textcolor[rgb]{ 1,  0,  0}{\textbf{0.669}} \\
		\hline
		MDNet~\cite{nam2016learning} & \textcolor[rgb]{ 0,  0,  1}{\textbf{0.565}} & \textcolor[rgb]{ 0,  0,  1}{\textbf{0.705}} & \textcolor[rgb]{ 0,  0,  1}{\textbf{0.606}} \\
		CFNet~\cite{valmadre2017end} & 0.533 & 0.654 & 0.578 \\
		SiamFC~\cite{bertinetto2016fully} & 0.533 & 0.663 & 0.571 \\
		ECO~\cite{danelljan2017eco}   & 0.492 & 0.618 & 0.554 \\
		CSRDCF~\cite{lukezic2017discriminative} & 0.48  & 0.622 & 0.534 \\
		SAMF~\cite{li2014scale}  & 0.477 & 0.598 & 0.504 \\
		ECO-HC~\cite{danelljan2017eco} & 0.476 & 0.608 & 0.541 \\
		Staple~\cite{bertinetto2016staple} & 0.470  & 0.603 & 0.528 \\
		Staple\_CA~\cite{mueller2017context} & 0.468 & 0.605 & 0.529 \\
		BACF~\cite{galoogahi2017learning}  & 0.461 & 0.580  & 0.523 \\
		\hline
	\end{tabular}%
	\label{tab:tn}%
\vspace{-1mm}
\end{table}%

\subsection{Ablation Experiment}

To validate the impact of different components, we conduct ablation experiments on LaSOT (Protocol \uppercase\expandafter{\romannumeral2})~\cite{fan2018lasot} and VOT-2017~\cite{kristan2017visual}.

\renewcommand\arraystretch{1.05}
\begin{table}[t]\small
	\centering
	\caption{Effect on the number of stages in C-RPN.}
	\begin{tabular}{r@{}c@{}c@{}c@{}}
		\hline
		\# Stages & \multicolumn{1}{l}{One stage} & \multicolumn{1}{l}{Two stages} & \multicolumn{1}{l}{Three stages} \\
		\hline \hline
		SUC on LaSOT  & 0.417 & 0.446 & 0.455 \\
		Speed on LaSOT & 48 fps  & 37 fps & 23 fps \\
		\hline
		EAO on VOT-2017   & 0.248 & 0.278 & 0.289 \\
		\hline
	\end{tabular}%
	\label{tab:stage}%
\vspace{-1mm}
\end{table}%

\renewcommand\arraystretch{1.05}
\begin{table}[t]\small
	\centering
	\caption{Effect on negative anchor filtering (NAF) in C-RPN.}
	\begin{tabular}{rcc}
		\hline
		& \multicolumn{1}{l}{C-RPN w/o NAF} & \multicolumn{1}{l}{C-RPN w/ NAF} \\
		\hline\hline
		SUC on LaSOT    & 0.439 &   0.455 \\
		\hline
		EAV on VOT-2017      & 0.282 &  0.289\\
		\hline
	\end{tabular}%
	\label{tab:easyneg}%
\vspace{-2mm}
\end{table}%

\renewcommand\arraystretch{1.05}
\begin{table}[t]\small
	\centering
	\caption{Effect on feature transfer block in C-RPN.}
	\begin{tabular}{rcc}
		\hline
		& \multicolumn{1}{l}{C-RPN w/o FTB} & \multicolumn{1}{l}{C-RPN w/ FTB} \\
		\hline\hline
		SUC on LaSOT    & 0.442 &   0.455 \\
		\hline
		EAV on VOT-2017      & 0.278 &  0.289\\
		\hline
	\end{tabular}%
	\label{tab:ftb}%
\vspace{-2mm}
\end{table}%

\vspace{0.1 em}
\noindent
{\bf Number of stages?} As shown in Tab.~\ref{tab:stage}, adding the second stage significantly improves one-stage baseline. The SUC on LaSOT is improved by 2.9\% from 0.417 to 0.446, and the EAO on VOT-2017 is increased by 3.5\% from 0.248 to 0.283. The third stage produces 0.9\% and 0.6\% improvements on LaSOT and VOT-2017, respectively. We observe that the improvement by the second stage is higher than that by the third stage. This suggests that most difficult background is handled in the second stage. Adding more stages may lead to further improvements, but also the computation (speed from 48 to 23 fps).

\vspace{0.1 em}
\noindent
{\bf Negative anchor filtering?} Filtering out the easy negatives aims to provide more balanced training samples for RPN in next stage. To show its effectiveness, we set threshold $\theta$ to 1 such that all refined anchors will be send to the next stage. Tab.~\ref{tab:easyneg} shows that removing negative anchors in C-RPN can improve the SUC on LaSOT by 1.6\% from 0.439 to 0.455, and the EAO on VOT-2017 by 0.7\% from 0.282 to 0.289, respectively, which evidences balanced training samples are crucial for training more discriminative RPN.

\vspace{0.1 em}
\noindent
{\bf Feature transfer block?} As demonstrated in Tab.~\ref{tab:ftb}, FTB improves the SUC on LaSOT by 1.3\% from 0.442 to 0.455 without losing much efficiency, and the EAO on VOT-2017 by 1.1\% from 0.278 to 0.289, validating the effectiveness of multi-level feature fusion in improving performance.

These studies show that each ingredient brings individual improvement, and all of them work together to produce the excellent tracking performance.

\vspace{-2mm}
\section{Conclusion}

In this paper, we propose a novel multi-stage framework C-RPN for tracking. Compared with previous arts, C-RPN demonstrates more robust performance in handling complex background such as similar distractors by performing hard negative sampling within a cascade architecture. In addition, the proposed FTB enables effective feature leverage across layers for more discriminative representation. Moreover, C-RPN progressively refines the target bounding box using multiple steps of regressions, leading to more accurate localization. In extensive experiments on six popular benchmarks, C-RPN consistently achieves the state-of-the-art results and runs in real-time.

\newpage

{\small
\bibliographystyle{ieee}
\bibliography{egbib}
}

\end{document}